\pdfoutput=1

\documentclass[11pt]{article}

\usepackage{EMNLP2022}

\usepackage{todonotes}
\presetkeys{todonotes}{color=green}{}

\usepackage{times}
\usepackage{latexsym}

\usepackage[T1]{fontenc}

\usepackage[utf8]{inputenc}

\usepackage{microtype}
\usepackage{booktabs}
\usepackage{multirow}
\usepackage{graphicx}
\usepackage{caption}
\usepackage{subcaption}
\usepackage{tabularx}
\usepackage{amssymb}
\usepackage{pifont}
\usepackage{adjustbox}
\usepackage{xcolor}

\newcommand{\cmark}{\ding{51}}%
\newcommand{\xmark}{\ding{55}}%
\newcommand{\filledsquare}{\ding{110}}
\newcommand{\filledcircle}{\ding{108}}
\newcommand{\filledtriangle}{\ding{115}}
\definecolor{green}{HTML}{3C9D4E}

\definecolor{mbert_color}{HTML}{648CCA}
\definecolor{canine_color}{HTML}{8DBDD5}
\definecolor{mt5_color}{HTML}{FFC889}
\definecolor{byt5_color}{HTML}{FE8C65}

\usepackage{inconsolata}

\usepackage{CJKutf8}

\title{A Multi-dimensional Evaluation of \\Tokenizer-free Multilingual Pretrained Models}

\author{
Jimin Sun\textsuperscript{1,2} \quad Patrick Fernandes\textsuperscript{1} \quad Xinyi Wang\textsuperscript{1} \quad Graham Neubig\textsuperscript{1}\\
\textsuperscript{1}Language Technologies Institute, Carnegie Mellon University \quad \textsuperscript{2}Kakao Enterprise\\
\texttt{\{jimins2,pfernand,xinyiw1,gneubig\}@cs.cmu.edu}}

\begin{document}

\maketitle
\begin{abstract}
Recent work on tokenizer-free multilingual pretrained models show promising results in improving cross-lingual transfer and reducing engineering overhead~\citep{Clark2022CaninePA,Xue2022ByT5TA}.
However, these works mainly focus on reporting accuracy on a limited set of tasks and data settings, placing less emphasis on other important factors when tuning and deploying the models in practice, such as memory usage, inference speed, and fine-tuning data robustness. We attempt to fill this gap by performing a comprehensive empirical comparison of multilingual tokenizer-free and subword-based models considering these various dimensions. Surprisingly, we find that subword-based models might still be the most practical choice in many settings, achieving better performance for lower inference latency and memory usage. Based on these results, we encourage future work in tokenizer-free methods to consider these factors when designing and evaluating new models.

\end{abstract}
\section{Introduction}

Several recent results \cite{Clark2022CaninePA,Xue2022ByT5TA} have excited the research community with the possibility of ``tokenizer-free'' models, character-level and byte-level models, as an alternative to more traditional subword-based models.
We, the authors of this paper, were also initially excited by these results -- the possibility of eschewing the two-step processing pipeline of subword segmentation and subword-based models would reduce the corresponding difficulties in cross-lingual transfer \cite{pmlr-v119-hu20b,maronikolakis-etal-2021-wine-v,rust-etal-2021-good,Wang2021-wy} or domain adaptation \cite{sato-etal-2020-vocabulary,liu-etal-2021-bridging} due to inconsistent subword units.
However, upon several attempts to apply tokenizer-free methods, our excitement was tempered upon realization of a number of practical difficulties in applying these methods.
This paper is a chronicle of some of the concerns we uncovered; we highlight some challenges with applying these models and propose best practices for future results reporting in this area.

Specifically, we perform experiments fine-tuning pretrained multilingual models, evaluating them with respect to (1) robustness to fine-tuning data settings, (2) data efficiency, and (3) inference time and memory consumption.
Based on these multiple dimensions, we come to the somewhat surprising conclusion that subword-based models might still be the most practical choice in most settings, as they are comparably robust to various fine-tuning data settings with a relatively low inference cost.

\begin{table*}[htp]
    \centering
    \resizebox{0.95\textwidth}{!}{%
    \begin{tabular}{l|llllllll}
        \toprule
         Model & Params &  Architecture & Enc. & Dec. & Tokenization & Downsample? & Pretrained corpus & Languages \\
        \midrule
         mBERT & 178M & Enc-only & 12 & - & Subword & \xmark{} & Wikipedia & 104 \\
         CANINE & 127M & Enc-only & 12 & - & Character & \cmark{} & Wikipedia & 104 \\
         \midrule
         mT5 (Small) & 300M & Enc-dec & 8 & 8 & Subword & \xmark{} & mC4 & 101 \\
         ByT5 (Small) & 300M & Enc-dec & 12 & 4& UTF-8 bytes & \xmark{}  & mC4 & 101 \\
         \bottomrule
    \end{tabular}%
    }
    \caption{Configuration of the pre-trained models used for experiments. From left to right: number of parameters, architecture, encoder depth, decoder depth, tokenization scheme, whether downsampling was used to reduce computation, pretrained corpus, number of languages covered during pretraining}
    \label{tab:models}
\end{table*}
\section{Tokenizer-free Multilingual Models}

While multilingual pretrained models~\citep{Devlin2019-sf,Lample2019CrosslingualLM,liu-etal-2020-multilingual-denoising,Chi2021InfoXLMAI,Xue2021mT5AM} have led to impressive performance for low-resource languages through cross-lingual transfer, the standard word representation in these models relies on subword segmentation~\citep{sennrich-etal-2016-neural,Kudo2018-subword-reg}. In a multilingual setting, subword tokenization can be sub-optimal as supporting hundreds of languages with various scripts and vocabulary causes segmentation mismatch between languages and over-segmentation in the lower-resourced languages~\citep{wang-etal-2020-extending,ebrahimi-kann-2021-adapt}. To alleviate this problem, recent works propose to remove the preprocessing step of subword segmentation by directly using characters or bytes as lexical units~\citep{Clark2022CaninePA, Xue2022ByT5TA}. 
\autoref{tab:models} presents an overview of the different tokenizer-free multilingual models with comparable subword models. Next, we briefly describe the two tokenizer-free models we consider in this work. 

\paragraph{CANINE}  
\cite{Clark2022CaninePA} is a character-level encoder-only model comparable with mBERT~\citep{Devlin2019-sf}. CANINE operates on character sequences and is pretrained using the masked language modeling~(MLM) objective. To compensate for the loss of computational efficiency due to increased sequence length, CANINE relies on convolution layers to down-sample the character sequence before feeding the representations to the transformer layers. 

The two model variants of CANINE -- CANINE-S and CANINE-C -- have the same architecture but slightly different pretraining strategies using subwords or characters. Our experiments found similar performance for both variants, so we only show the performance of CANINE-S, leaving the results for CANINE-C in \autoref{app:full_results}. 


\paragraph{ByT5} 
\cite{Xue2022ByT5TA} is an encoder-decoder transformer model similar to the mT5~\citep{Xue2021mT5AM} model. Both ByT5 and mT5 are pretrained on the multilingual Common Crawl (mC4) corpus\footnote{\url{https://www.tensorflow.org/datasets/catalog/c4\#c4multilingual}} using the span reconstruction objective proposed by \citet{JMLR:RaffelT5}. ByT5 operates on the raw UTF-8 bytes of the input without any downsampling, leading to a longer sequence length while having a much smaller vocabulary size than mT5. 

To keep the parameter count fixed between mT5 and ByT5, \citet{Xue2022ByT5TA} allocate the parameters saved from the embedding layer in ByT5 to additional encoder layers. Although a reasonable design choice, our results in \autoref{sec:eval} show that ByT5 suffers from a much higher inference cost due to a deeper encoder and longer sequence lengths in both the input and output.%
\footnote{We mainly consider the ByT5 small model to keep inference cost relatively constant and fit the model in the hardware that we easily have access to.} 

\paragraph{Discussion}
One significant advantage of multilingual pretrained models is that once the pretraining is complete, we can fine-tune the model to a variety of tasks and languages under different resource settings. In this context, it is also essential to consider the inference cost of the fine-tuned models. To this end, we conduct a multi-dimensional evaluation focusing on three aspects: robustness to fine-tuning data settings (\autoref{sec:robustness}), data efficiency (\autoref{sec:data_efficiency}), and inference cost (\autoref{sec:inf_cost}) to provide a better understanding of the practical applicability of tokenizer-free models.

\section{Experimental settings}
\begin{figure*}[ht]
\centering
\begin{subfigure}{0.92\textwidth}
  \centering
    \includegraphics[width=\textwidth]{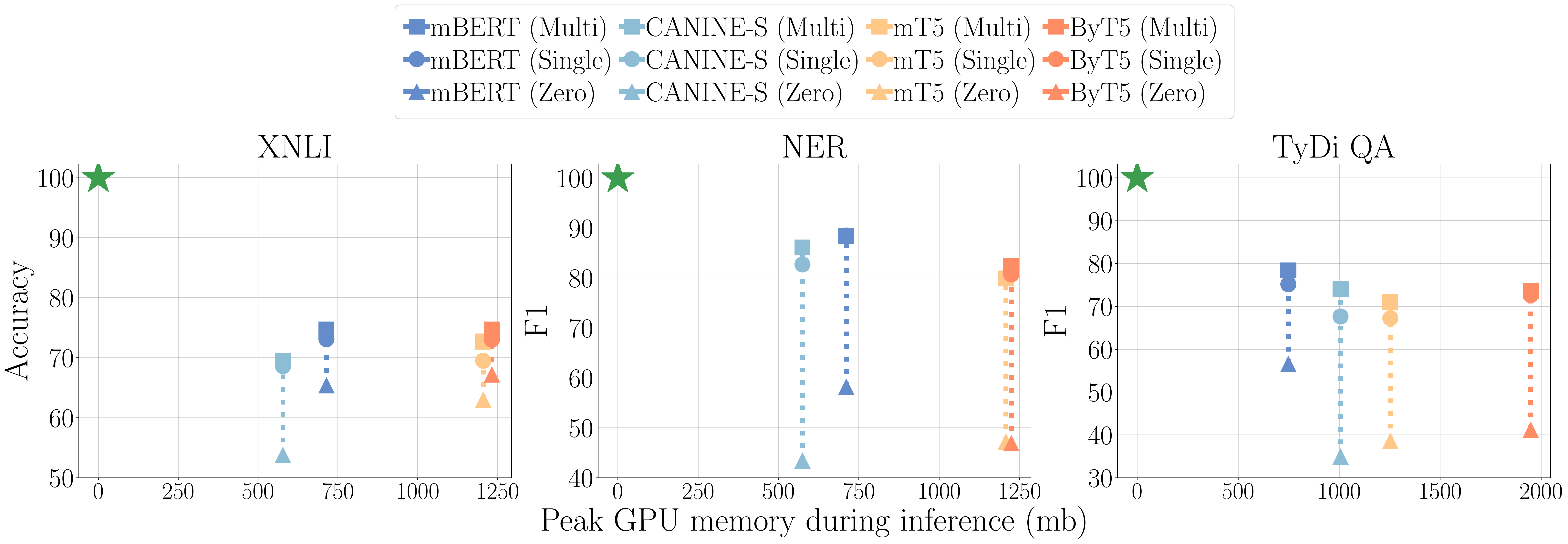}
\end{subfigure}\vspace{0em}\\
\begin{subfigure}{0.92\textwidth}
  \centering
    \includegraphics[width=\textwidth]{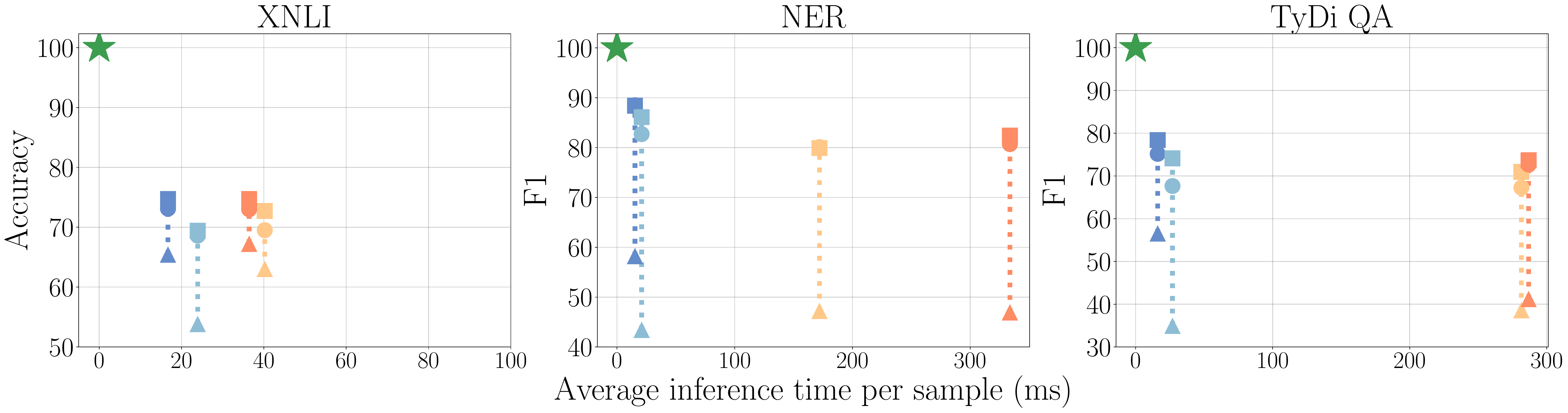}
\end{subfigure}

\caption{Average XNLI, NER, TyDi QA performance across languages when each model is fine-tuned with multilingual data (mBERT: \textcolor{mbert_color}{\filledsquare{}}, CANINE-S: \textcolor{canine_color}{\filledsquare{}}, mT5: \textcolor{mt5_color}{\filledsquare{}}, ByT5: \textcolor{byt5_color}{\filledsquare{}}), single-language data (mBERT: \textcolor{mbert_color}{\filledcircle{}}, CANINE-S: \textcolor{canine_color}{\filledcircle{}}, mT5: \textcolor{mt5_color}{\filledcircle{}}, ByT5: \textcolor{byt5_color}{\filledcircle{}}), or zero-shot transferred from English (mBERT: \textcolor{mbert_color}{\filledtriangle{}}, CANINE-S: \textcolor{canine_color}{\filledtriangle{}}, mT5: \textcolor{mt5_color}{\filledtriangle{}}, ByT5: \textcolor{byt5_color}{\filledtriangle{}}). The x-axis indicates the corresponding inference cost criteria (Top row: Peak GPU memory, Bottom row: Inference latency) in each task. We additionally use \textcolor{green}{\ding{77}} to mark the direction of the ideal cost and performance.\label{fig:main_result} 
}
\end{figure*}

We evaluate mBERT, CANINE, mT5, and ByT5 on three tasks adopted from the XTREME benchmark~\cite{pmlr-v119-hu20b}. 


\subsection{Tasks}
\paragraph{XNLI} The Cross-lingual Natural Language Inference \cite{DBLP:conf/emnlp/ConneauRLWBSS18} is a sequence-level classification task in which the model predicts whether the hypothesis sentence is an entailment, contradiction, or neutral given the premise sentence. The task is provided in 15 languages.
\paragraph{NER} Named Entity Recognition (NER) is a structured prediction task. We use the WikiAnn dataset \cite{DBLP:conf/acl/PanZMNKJ17}, which covers 282 languages. We select 20 languages (listed in \autoref{tab:result_per_language_ner}) based on linguistic diversity and the languages available in the other two tasks we consider. We use the train, dev, and test splits from \citet{DBLP:conf/acl/RahimiLC19}.
\paragraph{TyDi QA-GoldP} The Typologically Diverse Question Answering \cite{DBLP:journals/tacl/ClarkPNCGCK20} dataset is an extractive question answering benchmark in 11 languages. We use the gold passage version of the task (GoldP), which covers nine languages. 

\subsection{Details of Hardware and Measurements}
We use a single Tesla V100 (32GB) GPU for all experiments regarding inference cost measurements. To obtain the peak GPU memory and inference latency, we randomly select 100 samples from the English test set for each task and measure the average cost of predicting one example at a time.

\section{A Multi-dimensional Evaluation\label{sec:eval}}
In \autoref{fig:main_result}, We plot the performance of the models fine-tuned using three different data settings: 1) \textsc{Zero}: the model is fine-tuned with English training data and then evaluated on the multilingual test set; 2) \textsc{Single}: the model is fine-tuned individually on the task data in each language; 3) \textsc{Multi}: the model is jointly fine-tuned on the task data in all languages. The locations of the models on the x-axis are arranged from left to right based on increasing latency and memory cost during inference. 

Models located closer to the upper left corner of each plot are preferred because they achieve better performance on the test set while incurring lower inference costs. Interestingly, our multi-dimensional evaluation reveals that mBERT generally has the best performance and efficiency under most settings. Next, we discuss each evaluation dimension in detail.

\subsection{Robustness to fine-tuning data settings}
\label{sec:robustness}
\begin{figure*}[ht]
  \centering
  \resizebox{0.95\linewidth}{!}{
  \includegraphics[width=\textwidth]{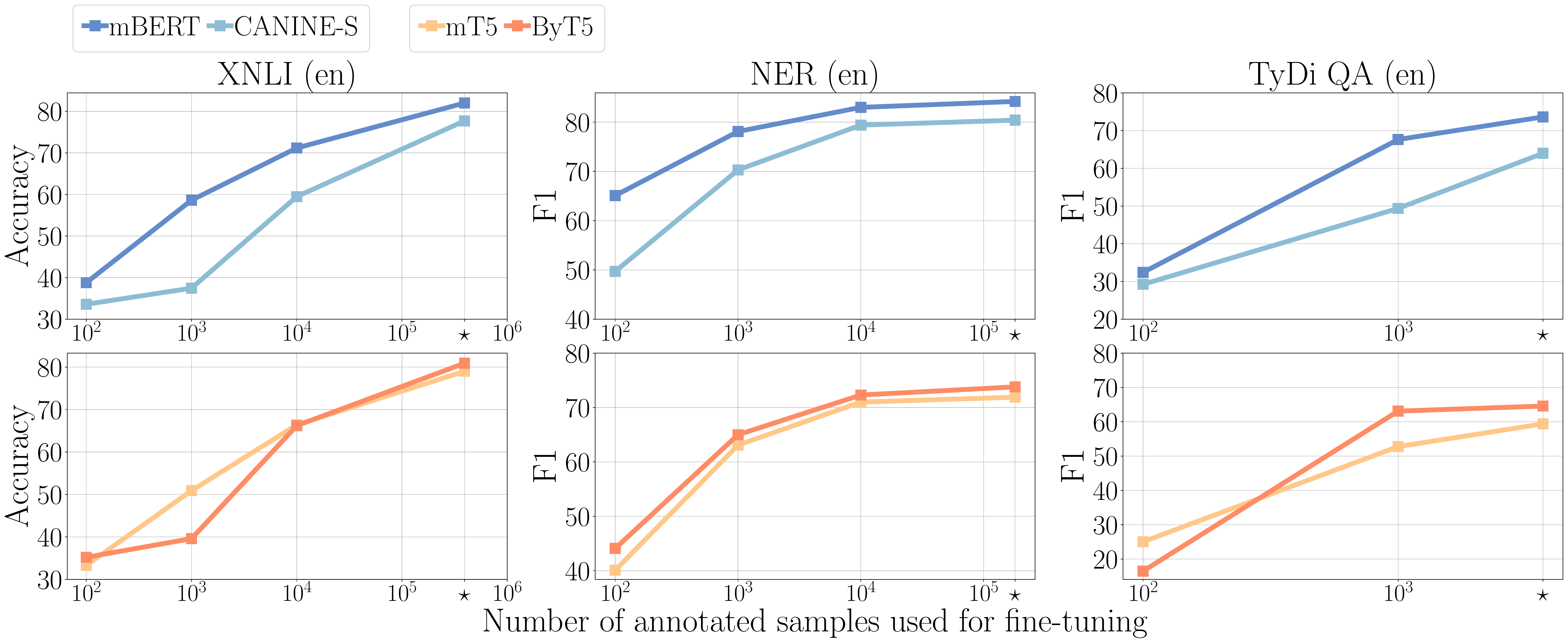}}
\caption{Performance of models fine-tuned with different numbers of labeled samples in the three tasks in English. The top row compares mBERT to the CANINE-S model, while the bottom row compares mT5 to ByT5. We also denote the full dataset size with $\star{}$ on the x-axis.}
\label{fig:data_efficiency}
\vspace{-1em}
\end{figure*}

\label{sec:overall_performance}
In \autoref{fig:main_result}, we plot the three performance numbers obtained under different fine-tuning data settings (\textsc{Multi}, \textsc{Single}, \textsc{Zero}) for each model. Overall, mBERT achieves the best accuracy for almost all three settings. Since \textsc{Multi} typically leads to the best performance, while \textsc{Zero} has the lowest performance, the length of the vertical line connecting the two points indicates how much the model performance fluctuates with different data settings. We find that CANINE generally has a larger performance gap under multiple data settings than mBERT, indicating that it is less robust to various fine-tuning settings. 

For mT5 and ByT5, we find that the two models perform similarly in terms of performance robustness. However, for downstream task scores, ByT5 significantly outperforms mT5 in most tasks and fine-tuning settings. We note that mT5-small might be somewhat penalized as most parameters are allocated on the embedding layer (85\% for mT5-Small vs 0.3\% for ByT5-Small). Also, given that the tasks concerned are not generation-heavy tasks, the extra depth on the encoder side (12 layers in ByT5-Small vs 8 layers in mT5-Small) might have favored ByT5 over mT5. However, when comparing the two encoder-decoder based models to mBERT, mBERT still achieves the best performance and robustness in all tasks and settings with the exception of XNLI Zero-shot.
\looseness=-1

\subsection{Fine-tuning data efficiency}
\label{sec:data_efficiency}
To further investigate how robustly each model performs under different resource conditions within a single language, we fine-tune the model with various task dataset sizes in English and present the results in \autoref{fig:data_efficiency}.\footnote{While we initially considered four languages (Arabic, English, Russian, Swahili) chosen based on the pretraining corpus size, we include other language results in \autoref{app:finetune_robustness} as they tend to exhibit similar performance trends.} Compared to mBERT, the performance of CANINE seems to degrade significantly as the amount of fine-tuning data decreases, particularly for XNLI and NER. To explain this phenomenon, we hypothesize that character-level models have the additional burden of learning to compose characters into semantically meaningful units, making it more difficult to generalize given a smaller amount of fine-tuning data. 
For mT5 and ByT5, we find that the two models perform comparably in smaller fine-tuning datasets while on larger fine-tuning data, ByT5 consistently outperforms mT5 on all three tasks.
\footnote{We suspect ByT5 matches/surpasses the performance of mT5 in many situations due to the extra depth in the encoder.} \looseness=-1

Finally, we observe that mBERT achieves the best performance on almost all tasks when the dataset is small, which is surprising, as we expect mT5/ByT5 models to generalize better in low-resource settings given their much larger pretraining corpus.

\subsection{Inference cost}
\label{sec:inf_cost}
Another key concern in utilizing pretrained models for downstream applications is the inference cost, such as memory consumption and latency. In the first row of \autoref{fig:main_result}, we visualize the models' performance to their peak GPU memory consumption. We see that the encoder-only models, mBERT and CANINE, require much less memory than the mT5 and ByT5 models. CANINE generally does not incur a higher memory cost than mBERT, as it has fewer parameters. However, ByT5 is more memory intensive compared to mT5, particularly for the TyDi QA task, as the task consists of relatively longer sequences, which is especially problematic for ByT5 as it has a much deeper encoder.

The plots in the second row of \autoref{fig:main_result} show the results based on the inference latency. mBERT has a slight advantage over CANINE, but the two models are generally comparable. However, mT5 and ByT5 have much lower inference speeds than mBERT. Because ByT5 has a smaller vocabulary, its inference speed is not much worse than mT5, despite having a deeper encoder and longer sequences to generate. In fact, ByT5 has a faster inference speed than mT5 for XNLI because the models only need to predict a single token.

In general, we find that mBERT and CANINE are both very accessible and efficient to use, whereas ByT5 would require more effort in tuning the batch size to fit into the GPU memory.


\section{Conclusion}
In this paper, we present a multi-dimensional evaluation of tokenizer-free multilingual pretrained models focusing on their robustness against fine-tuning data settings, data efficiency, and inference cost. Surprisingly, we find that mBERT might be the most practical choice so far, considering all the abovementioned factors. Despite our findings, tokenizer-free models still have a significant advantage in reducing engineering efforts and potentially increasing robustness to noisy and multilingual data. We believe more work should be done in developing \textit{efficient} tokenizer-free models that are \textit{robust} in various fine-tuning settings. Based on these results, we encourage the community to consider these criteria of practical applicability when developing and evaluating tokenizer-free pretrained models.

\section{Limitations}

This paper mainly covers three NLP tasks, focusing on the smaller-sized multilingual pretrained models. In future work, it would be interesting to run the multi-dimensional evaluation we suggest on a broader set of tasks and models. Although our results show that subword models are a more practical choice in some tasks, we note that other tasks or datasets may exist where tokenizer-free methods achieve better relative performance. For instance, tokenizer-free models have been reported to excel in word-level tasks, and noisy environments \cite{Xue2022ByT5TA}, and the conclusions we reached may be different in such settings. Moreover, we did not explore more complicated generation tasks like translation or summarization, where the difficulty in decoding and longer decode horizons could paint a different picture in a multi-dimensional evaluation.

\section*{Ethics Statement}
We hope our results encourage the community to consider the practical concerns of running large language models (LLMs) and designing tokenizer-free pretrained models. As the state-of-the-art LLMs are becoming more computationally extensive, it has become increasingly difficult for researchers and practitioners with less resources to utilize these models for downstream applications. We hope our multi-dimensional analysis can help researchers and practitioners with less computational resources decide which model to use in practice.

\section*{Acknowledgements}
We acknowledge Kakao Enterprise for providing the compute resources for this work. Additionally, we would like to thank Jon Clark for answering questions related to the CANINE model. This work was supported in part by grant \#2040926 from the National Science Foundation as well as the CMU-Portugal MAIA project.

\newpage
\bibliography{anthology}
\bibliographystyle{acl_natbib}

\newpage
\appendix
\label{sec:appendix}

\section{Tasks}

For all tasks and models, we refer to the original papers' codebase for hyperparameters.\footnote{\url{https://github.com/google-research/language/tree/master/language/canine}}\footnote{\url{https://github.com/google-research/multilingual-t5}}\footnote{\url{https://github.com/google-research/byt5}}

\paragraph{XNLI} For encoder-only models, the first token (\texttt{[CLS]}) is used to map the sentence representation to the label distribution. For encoder-decoder models, we generate the index of the label (e.g., \texttt{`0'}) directly.
\paragraph{NER} For encoder-decoder models, we follow the input-output format (e.g., input: \texttt{`tag: rick and morty are cool .'}, output: \texttt{`PER: rick \$\$ PER: morty'}) specified in the mT5 model's original codebase.
\section{Per language results}
\subsection{Main experiments (Zero, Single, Multi)}
\label{app:full_results}

\begin{table*}[t]
    \centering
    \resizebox{\textwidth}{!}{%
    \begin{tabular}{l|ccccccccccccccc|c}
    \toprule
    Model   & en & ar & bg & de & el & es & fr & hi & ru & sw & th & tr & ur & vi & zh & avg  \\
    \midrule
    \multicolumn{1}{l}{\textbf{Zero-shot (en)}} \vspace{0.2em} \\ 
        \hspace{2mm} mBERT & \textbf{82.0} & 64.1 & 67.5 & 70.4 & 65.5 & 73.7 & 72.8 & 59.3 & 67.4 & 50.2 & 53.2 & 60.2 & \textbf{57.5} & 68.7 & \textbf{68.1} & 65.4 \\
        \hspace{2mm} CANINE-S & 77.7 & 50.1 & 60.1 & 62.4 & 53.7 & 67.6 & 66.0 & 43.7 & 60.7 & 40.4 & 39.6 & 47.9 & 41.1 & 53.1 & 43.2 & 53.8 \\
        \hspace{2mm} CANINE-C & 77.1 & 53.1 & 61.4 & 63.5 & 58.3 & 68.5 & 66.4 & 47.7 & 63.3 & 41.0 & 39.2 & 48.8 & 44.4 & 53.4 & 39.1 & 55.0 \\
        \hspace{2mm} mT5-Small & 79.0 & 61.3 & 66.0 & 64.4 & 67.4 & 65.9 & 62.4 & \textbf{59.7} & 66.6 & 52.2 & \textbf{64.1} & 57.9 & 56.4 & 57.3 & 63.9 & 63.0 \\
        \hspace{2mm} ByT5-Small & 80.9 & \textbf{65.9} & \textbf{70.2} & \textbf{71.2} & \textbf{67.7} & \textbf{76.5} & \textbf{75.0} & 58.6 & \textbf{67.9} & \textbf{62.4} & 58.4 & \textbf{63.6} & 55.6 & \textbf{69.5} & 64.9 & \textbf{67.2} \\
    \midrule
    \multicolumn{1}{l}{\textbf{Single-language}} \vspace{0.2em} \\
        \hspace{2mm} mBERT & \textbf{82.0} & 70.6 & \textbf{76.2} & \textbf{76.6} & \textbf{75.1} & \textbf{77.7} & \textbf{77.4} & 67.0 & \textbf{74.8} & 66.3 & 65.7 & 72.5 & 62.9 & \textbf{75.9} & \textbf{76.4} & \textbf{73.1} \\
        \hspace{2mm} CANINE-S & 77.7 & 65.8 & 70.6 & 72.4 & 68.6 & 73.8 & 73.4 & 61.2 & 69.7 & 61.5 & 59.9 & 66.6 & 58.0 & 67.4 & 57.2 & 66.9 \\
        \hspace{2mm} CANINE-C & 77.1 & 66.2 & 71.1 & 72.0 & 69.8 & 72.8 & 72.6 & 62.3 & 68.6 & 60.8 & 57.1 & 65.7 & 58.2 & 67.3 & 60.0 & 66.8 \\
        \hspace{2mm} mT5-Small & 79.0 & 65.4 & 69.9 & 72.0 & 73.6 & 73.1 & 74.8 & 65.2 & 70.3 & 63.2 & 69.7 & 67.6 & 58.9 & 69.2 & 71.0 & 69.5 \\
        \hspace{2mm} ByT5-Small & 80.9 & \textbf{72.9} & 75.4 & 75.8 & \textbf{75.1} & \textbf{77.7} & 76.4 & \textbf{68.3} & 73.4 & \textbf{67.5} & \textbf{70.0} & \textbf{72.6} & \textbf{63.0} & 72.7 & 72.5 & 73.0 \\
    \midrule
    \multicolumn{1}{l}{\textbf{Multilingual}} \vspace{0.2em} \\ 
        \hspace{2mm} mBERT & \textbf{83.5} & 73.2 & 77.7 & \textbf{77.5} & 75.7 & \textbf{79.8} & \textbf{78.6} & \textbf{70.1} & \textbf{76.4} & 68.1 & 67.2 & \textbf{73.8} & 64.4 & \textbf{76.5} & \textbf{77.9} & \textbf{74.7} \\
        \hspace{2mm} CANINE-S & 79.1 & 69.7 & 75.0 & 74.9 & 72.5 & 76.3 & 75.3 & 65.2 & 73.0 & 65.0 & 62.3 & 68.9 & 64.1 & 71.3 & 65.6 & 70.5 \\
        \hspace{2mm} CANINE-C & 78.0 & 68.5 & 73.7 & 74.1 & 72.9 & 75.7 & 74.9 & 63.8 & 71.7 & 64.4 & 57.7 & 67.9 & 62.6 & 69.7 & 58.7 & 69.0 \\
        \hspace{2mm} mT5-Small & 79.9 & 70.3 & 74.7 & 74.9 & 74.4 & 76.5 & 75.5 & 67.7 & 73.7 & 68.1 & 71.2 & 71.9 & 65.4 & 72.4 & 73.2 & 72.7 \\
        \hspace{2mm} ByT5-Small & 81.0 & \textbf{73.3} & \textbf{77.8} & 76.5 & \textbf{76.5} & 78.5 & 77.2 & 70.0 & 75.6 & \textbf{71.3} & \textbf{71.4} & 73.6 & \textbf{68.3} & 75.7 & 74.1 & \textbf{74.7} \\
    \bottomrule
    \end{tabular}%
    }
    \caption{XNLI Performance (Accuracy)}
    \label{tab:result_per_language_xnli}
\end{table*}

\begin{table*}[t]
    \centering
    \resizebox{\textwidth}{!}{%
    \begin{tabular}{l|cccccccccccccccccccc|c}
    \toprule
    Model   & en & ar & bn & de & el & es & fi & fr & hi & id & ja & ko & ru & sw & ta & te & th & tr & ur & zh  & avg \\
    \midrule
    \multicolumn{1}{l}{\textbf{Zero-shot (en)}} \vspace{0.2em} \\ 
        \hspace{2mm} mBERT & \textbf{84.2} & 41.7 & \textbf{68.2} & \textbf{78.2} & \textbf{71.4} & 71.8 & \textbf{77.3} & \textbf{78.0} & \textbf{64.5} & \textbf{51.6} & 29.2 & \textbf{59.7} & \textbf{65.6} & \textbf{71.4} & \textbf{51.0} & \textbf{50.4} & 0.4 & \textbf{73.9} & 33.3 & \textbf{43.1} & \textbf{58.2} \\
        \hspace{2mm} CANINE-S & 80.8 & 29.6 & 49.6 & 70.7 & 63.5 & 66.4 & 66.7 & 74.1 & 41.1 & 47.3 & 0.5 & 29.3 & 57.7 & 59.8 & 28.4 & 19.7 & 0.1 & 55.8 & 22.0 & 5.4 & 43.4 \\
        \hspace{2mm} CANINE-C & 81.1 & 38.3 & 56.9 & 70.9 & 66.4 & 64.8 & 68.0 & 73.5 & 43.4 & 46.6 & 1.8 & 28.7 & 61.7 & 58.9 & 36.9 & 21.6 & 0.2 & 58.9 & 29.8 & 8.1 & 45.8 \\
        \hspace{2mm} mT5-Small & 71.9 & 32.9 & 56.6 & 67.1 & 42.3 & 70.0 & 65.1 & 75.3 & 56.2 & 45.3 & 25.5 & 23.9 & 36.9 & 49.0 & 38.0 & 35.9 & 3.6 & 58.7 & \textbf{58.7} & 31.3 & 47.2 \\
        \hspace{2mm} ByT5-Small & 73.8 & \textbf{45.9} & 61.5 & 70.7 & 67.7 & \textbf{79.4} & 67.1 & 77.4 & 57.1 & 46.2 & \textbf{31.3} & 26.2 & 46.7 & 60.2 & 31.9 & 27.9 & \textbf{9.6} & 23.3 & 1.3 & 32.8 & 46.9 \\
    \midrule
    \multicolumn{1}{l}{\textbf{Single-language}} \vspace{0.2em} \\
        \hspace{2mm} mBERT & \textbf{84.2} & \textbf{89.6} & \textbf{96.1} & \textbf{90.3} & \textbf{91.4} & \textbf{92.5} & \textbf{92.2} & \textbf{91.2} & \textbf{91.2} & \textbf{93.6} & \textbf{74.4} & \textbf{88.8} & \textbf{89.4} & \textbf{90.0} & \textbf{86.5} & \textbf{80.4} & \textbf{76.2} & \textbf{93.2} & \textbf{95.7} & \textbf{83.1} & \textbf{88.5} \\
        \hspace{2mm} CANINE-S & 80.8 & 84.9 & 92.9 & 88.0 & 88.6 & 89.7 & 89.1 & 88.9 & 84.9 & 90.9 & 63.3 & 81.6 & 86.5 & 87.7 & 81.0 & 49.9 & 70.5 & 90.9 & 91.0 & 73.2 & 82.7 \\
        \hspace{2mm} CANINE-C & 81.1 & 85.1 & 93.5 & 87.5 & 89.1 & 89.8 & 88.4 & 88.4 & 84.3 & 90.6 & 60.2 & 79.5 & 87.3 & 86.5 & 79.6 & 43.0 & 74.0 & 90.6 & 92.4 & 68.9 & 82.0 \\
        \hspace{2mm} mT5-Small & 71.9 & 86.5 & 86.6 & 83.7 & 83.8 & 88.0 & 87.8 & 86.7 & 85.5 & 85.3 & 65.9 & 80.2 & 64.0 & 71.0 & 82.6 & 74.5 & 64.6 & 86.3 & 93.0 & 75.1 & 80.1 \\
        \hspace{2mm} ByT5-Small & 73.8 & 85.3 & 88.3 & 82.4 & 87.6 & 86.6 & 86.4 & 84.7 & 83.0 & 84.5 & 69.9 & 83.2 & 62.6 & 84.5 & 80.3 & 69.1 & 74.5 & 83.4 & 90.5 & 73.2 & 80.7 \\
    \midrule
    \multicolumn{1}{l}{\textbf{Multilingual}} \vspace{0.2em} \\
        \hspace{2mm} mBERT & \textbf{85.4} & \textbf{89.6} & \textbf{95.9} & \textbf{89.8} & \textbf{91.3} & \textbf{92.9} & \textbf{92.0} & \textbf{91.2} & \textbf{89.3} & \textbf{93.4} & \textbf{74.9} & \textbf{88.1} & \textbf{89.2} & \textbf{90.9} & \textbf{86.0} & \textbf{80.6} & 76.5 & \textbf{93.1} & \textbf{95.5} & \textbf{82.3} & \textbf{88.4} \\
        \hspace{2mm} CANINE-S & 84.1 & 88.0 & 94.7 & 89.3 & 90.7 & 92.1 & 91.1 & 90.9 & 85.8 & 92.8 & 69.3 & 83.8 & 88.8 & 89.6 & 81.7 & 71.3 & 76.2 & 92.4 & 94.0 & 75.7 & 86.1 \\
        \hspace{2mm} CANINE-C & 84.1 & 87.8 & 95.6 & 89.2 & 91.1 & 92.5 & 90.7 & 90.9 & 88.2 & 92.6 & 67.9 & 81.5 & 88.9 & 90.0 & 81.6 & 69.5 & \textbf{77.7} & 92.0 & 93.7 & 72.1 & 85.9 \\
        \hspace{2mm} mT5-Small & 72.5 & 86.8 & 84.5 & 84.8 & 83.4 & 88.7 & 88.3 & 87.7 & 83.6 & 87.2 & 70.1 & 83.1 & 64.8 & 72.3 & 82.3 & 69.8 & 67.8 & 86.9 & 92.4 & 76.5 & 80.7 \\
        \hspace{2mm} ByT5-Small & 73.5 & 87.7 & 88.4 & 86.1 & 88.7 & 90.3 & 89.9 & 89.3 & 84.7 & 87.3 & 70.3 & 83.8 & 66.3 & 84.3 & 81.8 & 78.0 & 72.6 & 88.6 & 92.6 & 76.5 & 83.0 \\
    \bottomrule
    \end{tabular}%
    }
    \caption{NER Performance (F1)}
    \label{tab:result_per_language_ner}
\end{table*}

\begin{table*}[t]
    \centering
    \begin{tabular}{l|ccccccccc|c}
    \toprule
    Model   & en & ar & bn & fi & id & ko & ru & sw & te & avg   \\
    \midrule
    \multicolumn{1}{l}{\textbf{Zero-shot (en)}} \vspace{0.2em} \\
        \hspace{2mm} mBERT & \textbf{73.64} & \textbf{60.11} & \textbf{45.1} & \textbf{57.63} & \textbf{63.78} & \textbf{52.16} & \textbf{57.52} & \textbf{56.51} & \textbf{42.15} & \textbf{56.51} \\
        \hspace{2mm} CANINE-S & 64.78 & 44.85 & 20.13 & 39.73 & 43.78 & 13.67 & 44.49 & 30.64 & 31.59 & 37.07 \\
        \hspace{2mm} CANINE-C & 63.96 & 42.19 & 22.05 & 43.13 & 36.87 & 17.44 & 42.02 & 33.3 & 30.51 & 36.83 \\
        \hspace{2mm} mT5-Small & 59.39 & 43.25 & 22.51 & 44.27 & 48.7 & 22.05 & 44.85 & 33.08 & 28.77 & 38.54 \\
        \hspace{2mm} ByT5-Small & 64.58 & 56.4 & 15.86 & 51.91 & 55.85 & 22.21 & 54.11 & 35.44 & 31.43 & 43.09 \\
    \midrule
    \multicolumn{1}{l}{\textbf{Single-language}} \vspace{0.2em} \\
        \hspace{2mm} mBERT & \textbf{73.64} & \textbf{79.86} & \textbf{70.78} & \textbf{76.08} & 79.93 & \textbf{62.76} & \textbf{72.48} & \textbf{79.81} & 81.21 & \textbf{75.17} \\
        \hspace{2mm} CANINE-S & 64.78 & 79.2 & 55.81 & 70.13 & 70.0 & 49.53 & 67.15 & 71.26 & 81.75 & 67.73 \\
        \hspace{2mm} CANINE-C & 63.96 & 77.79 & 50.92 & 67.28 & 66.26 & 49.84 & 66.49 & 71.39 & 82.78 & 66.3 \\
        \hspace{2mm} mT5-Small & 59.39 & 73.07 & 67.92 & 65.33 & 73.65 & 54.93 & 66.13 & 71.49 & 80.93 & 68.09 \\
        \hspace{2mm} ByT5-Small & 64.58 & 75.82 & 69.91 & 71.98 & \textbf{80.55} & 58.65 & 71.09 & 78.81 & \textbf{85.39} & 72.97 \\
    \midrule
    \multicolumn{1}{l}{\textbf{Multilingual}} \vspace{0.2em} \\
        \hspace{2mm} mBERT & \textbf{76.02} & \textbf{81.49} & 72.86 & \textbf{80.41} & \textbf{84.87} & \textbf{67.09} & \textbf{74.45} & \textbf{82.42} & 83.52 & \textbf{78.13} \\
        \hspace{2mm} CANINE-S & 71.55 & 80.53 & 67.24 & 75.42 & 78.44 & 61.25 & 71.75 & 77.43 & 83.53 & 74.13 \\
        \hspace{2mm} CANINE-C & 71.56 & 80.74 & 62.6 & 74.21 & 76.28 & 65.79 & 72.66 & 79.71 & 84.43 & 74.22 \\
        \hspace{2mm} mT5-Small & 64.39 & 75.34 & \textbf{76.89} & 70.01 & 76.73 & 59.24 & 67.86 & 76.62 & 81.35 & 72.05 \\
        \hspace{2mm} ByT5-Small & 69.42 & 75.86 & 70.9 & 74.52 & 79.78 & 60.62 & 73.01 & 80.32 & \textbf{85.93} & 74.48 \\
    \bottomrule
    \end{tabular}
    \caption{TyDi QA-GoldP Performance (F1)}
    \label{tab:result_per_language_tydi}
\end{table*}
XNLI: \autoref{tab:result_per_language_xnli}, NER: \autoref{tab:result_per_language_ner}, TyDiQA-GoldP: \autoref{tab:result_per_language_tydi}
\subsection{Single language fine-tuning robustness}
\label{app:finetune_robustness}
\begin{figure*}[ht]
  \centering
  \includegraphics[width=\textwidth]{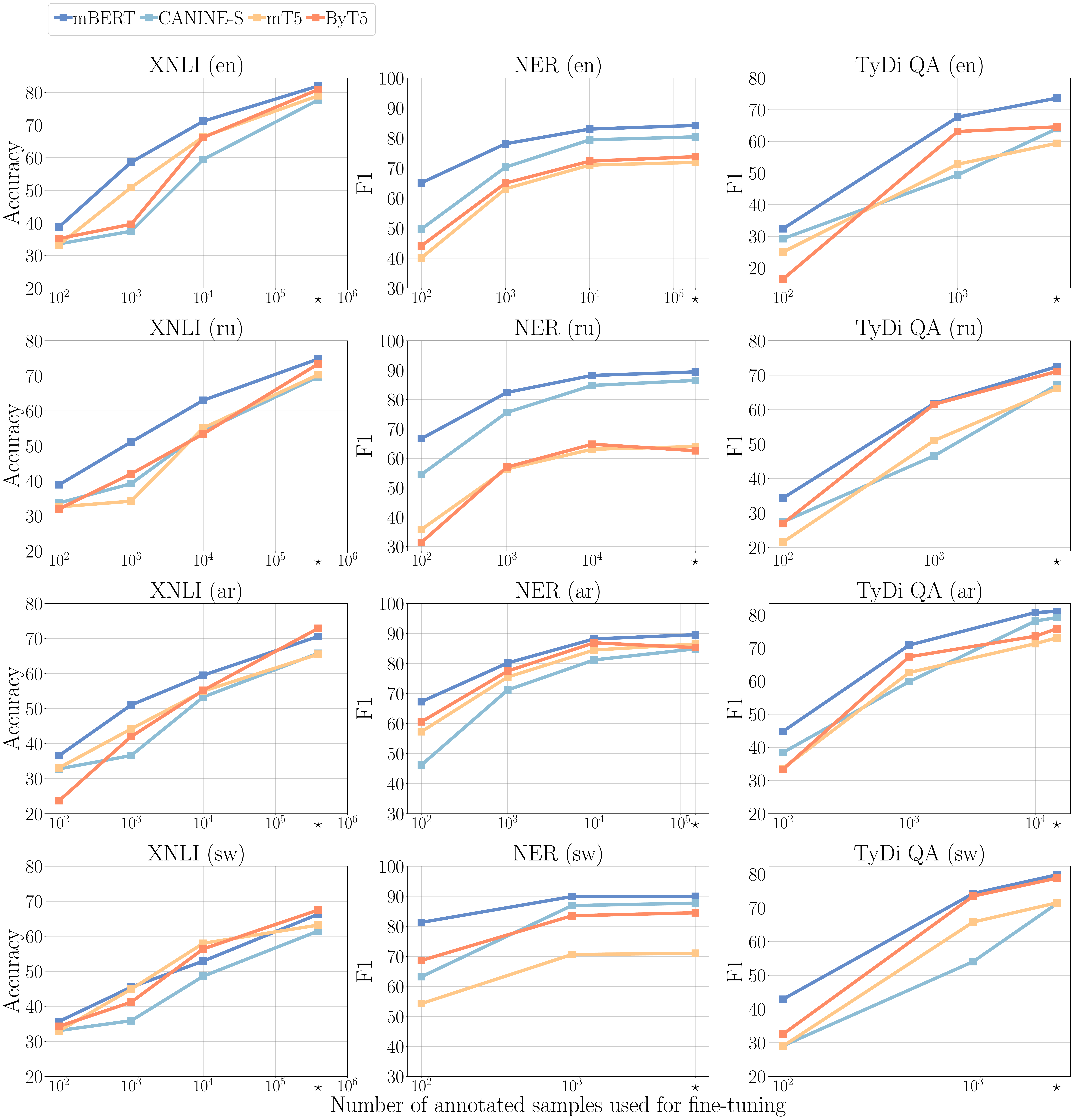}
\caption{Performance of models fine-tuned with different numbers of labeled samples in the three tasks in English (en), Russian (ru), Arabic (ar), and Swahili (sw). We also denote the full dataset size with $\star{}$ on the x-axis.}
\label{fig:data_efficiency_all_langs}
\end{figure*}

\autoref{fig:data_efficiency_all_langs}
\end{document}